\documentclass[11pt]{article}

\usepackage[preprint]{acl}
\usepackage{booktabs}
\usepackage{graphicx}
\usepackage{times}
\usepackage{latexsym}
\usepackage{CJKutf8}
\usepackage[table]{xcolor} 
\usepackage[T1]{fontenc}
\usepackage{algorithm}
\usepackage{algorithmic}
\usepackage[utf8]{inputenc}

\usepackage{microtype}
\usepackage{multirow} 
\usepackage{inconsolata}

\usepackage{graphicx}
\usepackage{enumitem}
\title{CFMS: Towards Explainable and Fine-Grained \\ Chinese Multimodal Sarcasm Detection Benchmark}

\author{Junzhao Zhang \and Hsiu-Yuan Huang \and Chenming Tang \and Yutong Yang \and Yunfang Wu$^*$ \\
National Key Laboratory for Multimedia Information Processing, Peking University \\
School of Computer Science, Peking University \\
School of Software and Microelectronics, Peking University \\
\texttt{\{zhangjunzhao, huang.hsiuyuan, tangchenming\}@stu.pku.edu.cn, \{yytpku, wuyf$^*$\}@pku.edu.cn}
}

\usepackage{amsmath}   
\usepackage{amsfonts}  
\usepackage{amssymb}   

\begin{document}
\maketitle
\begin{abstract}
Multimodal sarcasm detection has recently garnered significant attention. However, existing benchmarks suffer from coarse-grained annotations and limited cultural coverage, which hinder research into fine-grained semantic understanding. To address this, we construct \textbf{CFMS}, the first fine-grained multimodal sarcasm dataset tailored for Chinese social media. It comprises 2,796 high-quality image-text pairs and provides a triple-level annotation framework: sarcasm identification, target recognition, and explanation generation. We find that the fine-grained explanation annotations effectively guide AI in generating images with explicit sarcastic intent. Furthermore, we curate a high-consistency parallel Chinese-English metaphor subset (200 entries each), revealing significant limitations of current models in metaphoric reasoning. To overcome the constraints of traditional retrieval methods, we propose a Reinforcement Learning-augmented In-Context Learning strategy (\textbf{PGDS}) to dynamically optimize exemplar selection. Extensive experiments demonstrate that CFMS provides a solid foundation for building reliable multimodal sarcasm understanding systems, and the PGDS method significantly outperforms existing baselines on key tasks. Our data and code are available at \url{https://anonymous.4open.science/r/CFMS-E8F9}.
\end{abstract}

\section{Introduction}
\begin{CJK*}{UTF8}{gbsn}
\begin{figure}[t]
    \centering
    \includegraphics[width=0.48\linewidth]{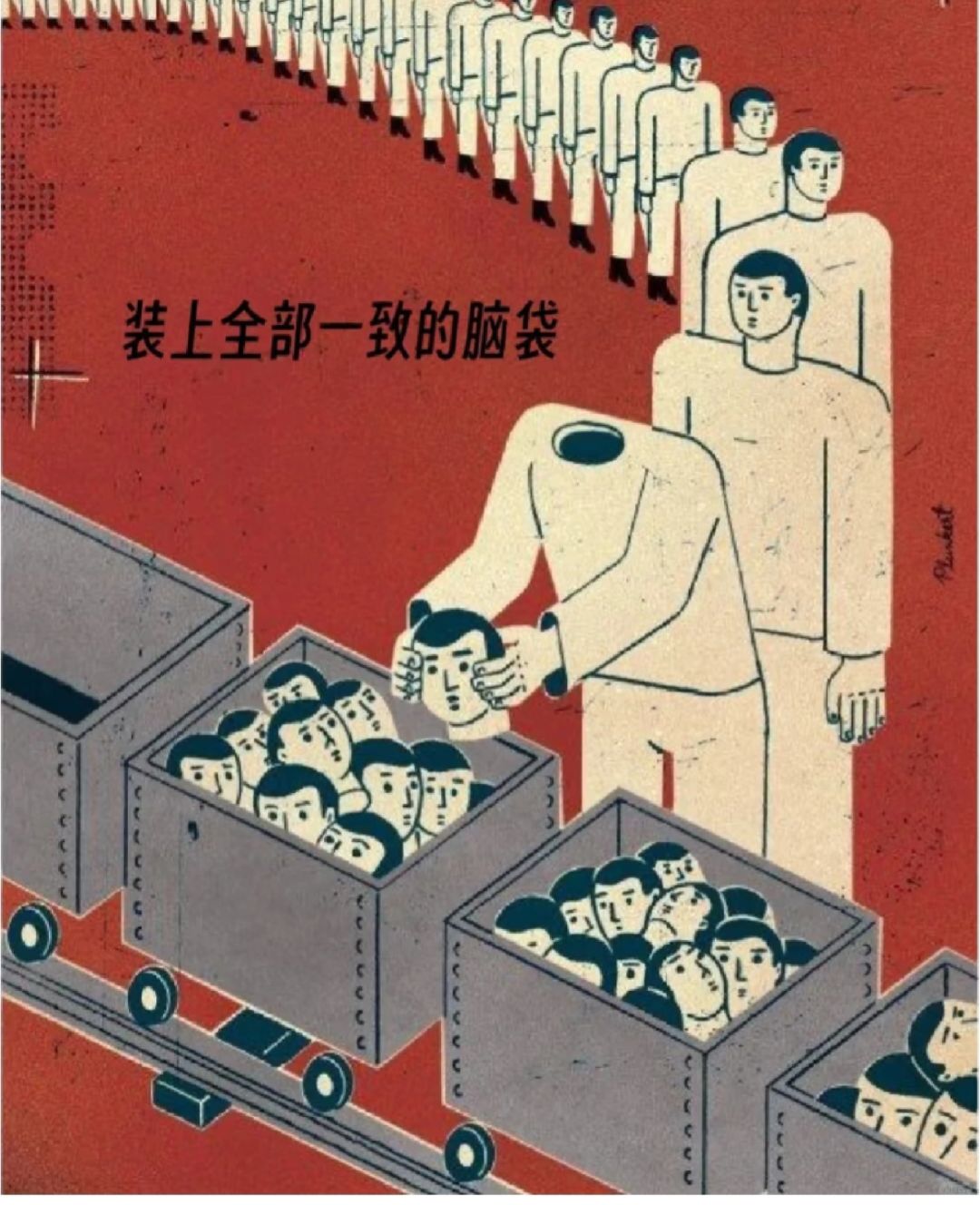} 
    \caption{A representative instance of Chinese multimodal sarcasm from the CFMS dataset. The neutral text ``装上全部一致的脑袋'' (lit. ``Install fully consistent heads'') contrasts with the visual of a conveyor belt, satirizing ideological conformity via culturally-grounded semantic inversion.}
    \label{fig:conveyor_belt}
\end{figure}

Sarcasm is a sophisticated linguistic phenomenon where literal meanings deviate from true intents to convey mockery or criticism \citep{camp2012sarcasm}. In the multimodal era, sarcasm frequently manifests through the conflict between text and imagery \citep{schifanella2016multimodal}. As illustrated in Figure \ref{fig:conveyor_belt}, the phrase ``Install fully consistent heads'' appears superficially neutral, yet its pairing with a mechanized conveyor belt reveals a deep-seated social critique. Resolving such \textbf{culturally-grounded semantic inversion} constitutes a primary challenge for automated systems.

While multimodal sarcasm detection (MSD) has progressed with benchmarks like MMSD2.0 \citep{qin2023multi}, existing research still faces three primary limitations:
(1) \textbf{Cultural Bias:} Most datasets are English-centric, failing to capture unique Chinese sarcastic forms like ``\begin{CJK}{UTF8}{gbsn}阴阳怪气\end{CJK}'' (passive-aggression). 
(2) \textbf{Coarse-grained Annotation:} Binary classification tasks often ignore the \textit{target} and \textit{construction mechanism}, limiting analytical depth. 
(3) \textbf{Shallow Heuristics:} Models frequently rely on surface cues (e.g., hashtags) rather than integrated multimodal reasoning \citep{cai2019multi}.

To address these gaps, we propose \textbf{CFMS} (Chinese Fine-grained Multimodal Sarcasm), a benchmark oriented toward real-world Chinese social media. CFMS shifts the focus from classification to interpretation via a \textbf{triple-level annotation} framework: \textit{sarcasm identification}, \textit{target recognition}, and \textit{explanation generation}. 

Furthermore, we explore the \textbf{Transferability} of CFMS beyond detection. Inspired by the synergy between sarcasm and metaphor in persuasive communication \cite{van2005puns}, we demonstrate that our reason-level explanations can serve as \textit{structured instruction signals} for AI-Generated Content (AIGC). In domains like creative advertising, CFMS enables Text-to-Image models to manifest implicit rhetorical intents that surface-level prompts fail to evoke \cite{oppenlaender2024taxonomy}, bridging the gap between understanding irony and generating it.

Regarding methodology, we evaluate various MLLMs and propose \textbf{Policy-Guided Demonstration Selection (PGDS)}. PGDS is a reinforcement-learned strategy that dynamically optimizes the selection of in-context exemplars based on task-level feedback, effectively improving fine-grained sarcasm understanding without modifying model parameters.

Our contributions are as follows:
\begin{itemize}[leftmargin=*,noitemsep,topsep=2pt]
    \item We provide an in-depth analysis of the limitations in current MSD benchmarks regarding annotation granularity and cultural coverage.
    \item We construct \textbf{CFMS}, the first fine-grained Chinese multimodal sarcasm benchmark with a triple-annotation system to facilitate interpretable sarcasm understanding.
    \item We propose \textbf{PGDS}, a reinforcement-learned strategy for dynamic exemplar selection that significantly enhances model performance in complex scenarios.
    \item We validate the \textbf{transferability} of fine-grained sarcasm explanations, demonstrating their value in guiding creative AI content generation.
\end{itemize}
\end{CJK*} 
\section{Related Work}

Multimodal sarcasm detection originated with the work of \citet{schifanella2016multimodal}, who first attempted to identify sarcasm by concatenating visual and textual features. \citet{cai2019multi} released the MMSD dataset, which became one of the most widely used benchmarks in the field. However, this dataset has been noted for containing significant spurious cues and annotation inconsistencies.

To address these reliability issues, \citet{qin2023multi} introduced MMSD2.0, which improved data quality by removing noisy samples and re-annotating unreasonable instances. From a different perspective, \citet{castro2019towards} constructed the MUStARD dataset using video clips from popular TV shows such as \textit{Friends} and \textit{The Big Bang Theory}, providing richer contextual information. Nevertheless, MUStARD is limited by its small scale and single cultural background.

In terms of methodology, early approaches evolved from simple feature concatenation to hierarchical fusion models based on attention mechanisms. \citet{liang2022multi} proposed a cross-modal graph convolutional network to model atomic-level and composition-level consistencies across modalities. Recently, \citet{yue2024sarcnet} introduced SarcNet, the first bilingual multimodal sarcasm dataset; however, it does not yet address fine-grained tasks such as target recognition or explanation generation. In contrast, CFMS focuses on the Chinese social media context and is the first to incorporate target recognition and explanation generation, pushing the field towards interpretability. Notably, a recent survey by \citet{farabi2024survey} emphasizes that high-quality dataset construction and fine-grained task design are paramount for future research—a principle that our work strictly adheres to.

\section{Dataset Construction}
\subsection{Preliminaries: Cognitive Framework for Sarcasm}
\label{subsec:preliminary}
The design of the CFMS dataset is grounded in the cognitive mechanisms of sarcasm. Human comprehension of sarcasm stems from complex processes involving semantic conflict, pragmatic reasoning, and cultural background knowledge \citep{gibbs2017metaphor}. While MLLMs have shown promise in general sentiment analysis, they often struggle with scenarios relying on deep semantic conflict reasoning—areas where humans maintain a significant advantage by evaluating the ``sarcastic effect'' \citep{lu2023seeing}. 

To bridge this gap, we adopt a structured expression of sarcasm explanation. Unlike previous benchmarks that focus solely on identification, we follow the insights of \citet{konstantinidou2025navigating} to capture semantic cues such as visual anomalies and emotional misalignment. In CFMS, we translate these human cognitive processes into computable, fine-grained annotations. This framework not only provides an interpretable path for models but also acts as structured guidance for generative tasks, moving from ``understanding'' to ``generating'' sarcastic intent.
\subsection{Sarcasm Data Collection}
We select a major Chinese social media platform as our primary data source, whose young user base frequently employs domain-specific sarcastic expressions like ``\begin{CJK}{UTF8}{gbsn}阴阳怪气\end{CJK}'' (passive-aggressive speech) and ``\begin{CJK}{UTF8}{gbsn}内涵\end{CJK}'' (veiled allusions). This provides a sharp contrast to existing English-centric datasets like MMSD \citep{cai2019multi} and MUStARD \citep{castro2019towards}, allowing us to capture the unique sarcastic culture of Chinese social media.

Regarding data acquisition, we adopt a directed crawling strategy. Unlike the Twitter API method used in MMSD, we utilize a distributed crawler framework to perform keyword-based searches using terms such as ``sarcasm,'' ``irony,'' ``connotation images,'' and ``passive-aggression,'' obtaining approximately 10,000 raw entries. To resolve the modality misalignment issue (where one text is paired with multiple images), we employ PaddleOCR to extract embedded text within images and perform semantic fusion with external captions, constructing strongly correlated ``image-text'' pairs.

Then, A triple-stage purification pipeline was implemented:
(1) \textbf{pHash Deduplication:} Removing duplicate samples with a similarity score $>90\%$.
(2) \textbf{Commercial Filter:} Filtering advertisements by detecting watermark area ($>15\%$) and commercial logos.
(3) \textbf{Low-Resolution Filter:} Discarding images with pixels $<512 \times 512$. 
After purification, approximately 3,500 candidate samples remained.

\begin{figure}[h]
    \centering
    \includegraphics[width=\columnwidth]{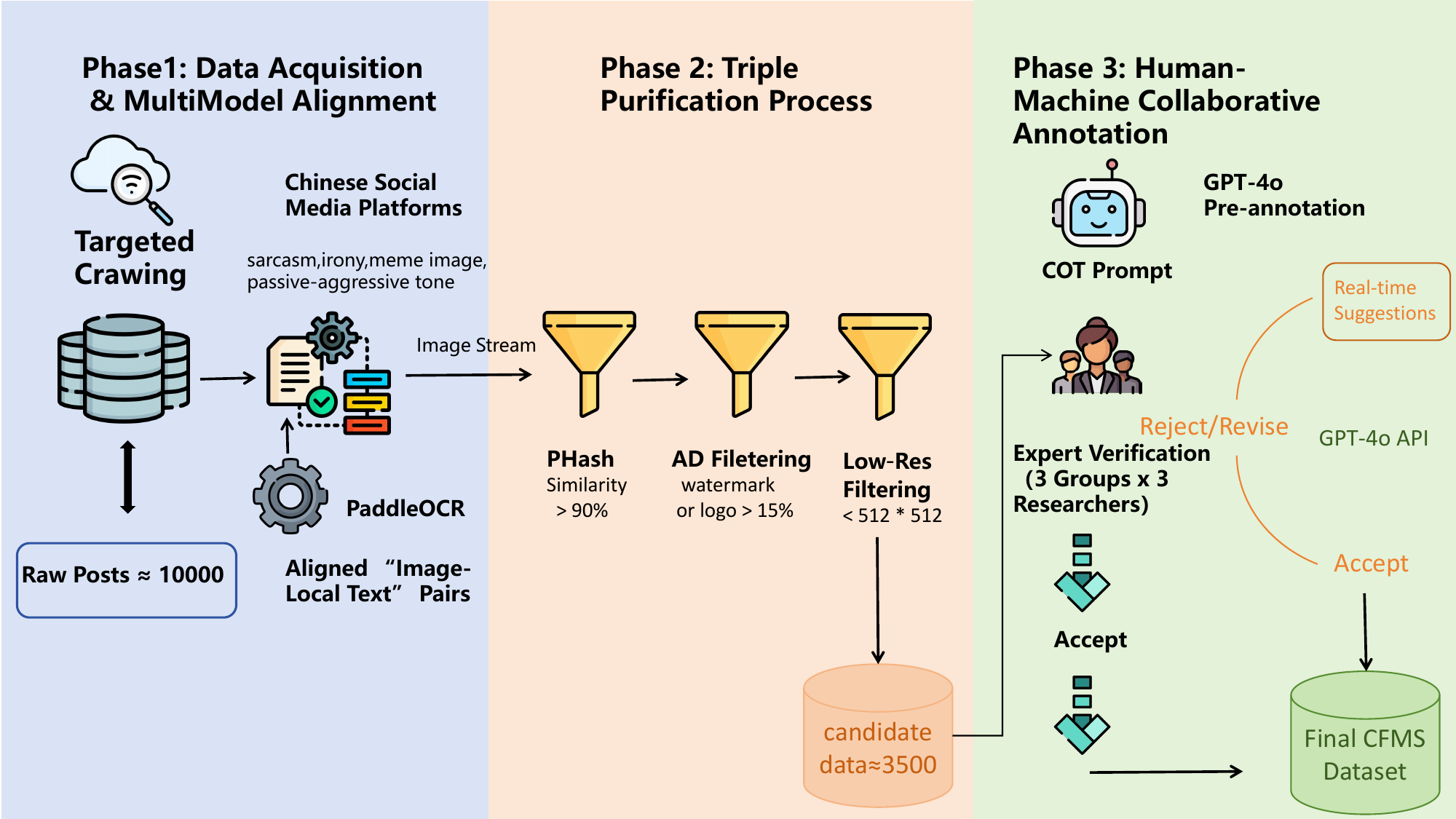} 
    \caption{The human-in-the-loop annotation pipeline for the CFMS dataset, integrating GPT-4o Pre-annotation, Human Verification, and Interactive Error Correction.}
    \label{fig:workflow}
\end{figure}

For the annotation process, we designed an ``LLM Pre-annotation + Human Verification'' pipeline inspired by the \textbf{human-in-the-loop (HITL)} paradigm \citep{munro2021human}, which leverages the synergy between automated efficiency and human expertise \citep{wang2021want, gilardi2023chatgpt}. The workflow (illustrated in Figure \ref{fig:workflow}) follows three steps: (1) \textbf{GPT-4o Pre-annotation} using a structured prompt (logic: ``Explicit Element Recording $\rightarrow$ Conflict Detection $\rightarrow$ Counter-Hypothesis Verification''); (2) \textbf{Human Verification} by three groups of graduate students in computational linguistics; (3) \textbf{Interactive Error Correction} via an integrated GPT-4o API for real-time feedback.
The custom-developed annotation interface is illustrated in Appendix \ref{subsec:app_ui}.

The annotation achieved a Kappa coefficient of 0.69 (``substantial agreement''), comparable to MUStARD \citep{castro2019towards}. The BLEU-4 consistency for the explanation annotation reached 0.68, while the Kappa was 0.63, indicating high annotation reliability. The final CFMS dataset contains 2,796 samples. The text length statistics are summarized in Table \ref{tab:len_stat}, and the dataset split is detailed in Table \ref{tab:distribution}.

\begin{table}[ht]
\centering
\small
\begin{tabular}{lc}
\toprule
\textbf{Category} & \textbf{Avg. Length (Chars)} \\ \midrule
Text Context      & 18.17 \\
Sarcasm Target    & 5.44  \\
Sarcasm Explanation & 61.69 \\ \bottomrule
\end{tabular}
\caption{Text length distribution in CFMS (N=2,796).}
\label{tab:len_stat}
\end{table}

\begin{table}[ht]
\centering
\small
\begin{tabular}{lrrr}
\toprule
\textbf{Split} & \textbf{Positive} & \textbf{Negative} & \textbf{Total} \\ \midrule
Training       & 872               & 1,084             & 1,956          \\
Validation     & 187               & 233               & 420            \\
Test           & 187               & 233               & 420            \\ \midrule
\textbf{Total} & \textbf{1,246}    & \textbf{1,550}    & \textbf{2,796} \\
\bottomrule
\end{tabular}
\caption{Statistics and distribution of the CFMS dataset.}
\label{tab:distribution}
\end{table}

\begin{figure}[h]
    \centering
    \includegraphics[width=\columnwidth]{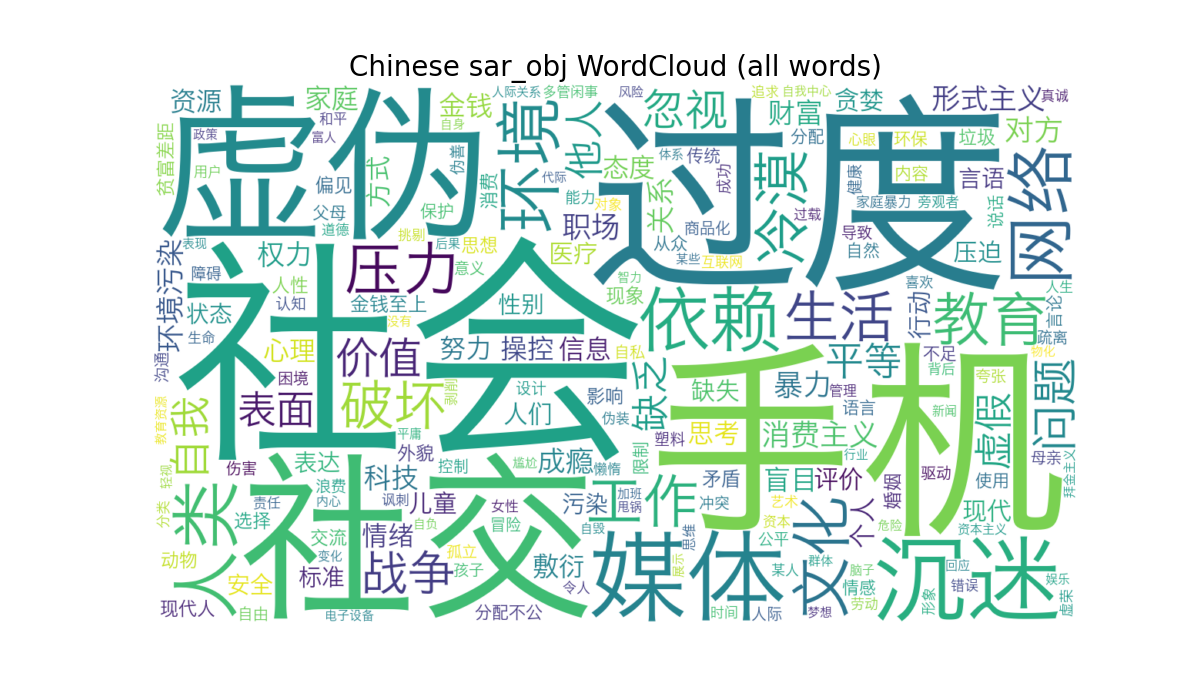} 
    \caption{Word cloud of sarcasm targets in the Chinese CFMS dataset. Larger fonts indicate more frequently identified targets, such as ``social phenomena'' and ``individual behavior.''}
    \label{fig:wordcloud_zh}
\end{figure}

As shown in the sarcasm target word cloud (Figure \ref{fig:wordcloud_zh}), targets are categorized into five types: \textit{Social Phenomena} (41\%), \textit{Individual Behavior} (23\%), \textit{Interpersonal Relations} (17\%), \textit{Institutional Rules} (12\%), and \textit{Others} (7\%). 
Table \ref{tab:cfms_cases} presents three representative samples with their corresponding fine-grained annotations.

\begin{CJK*}{UTF8}{gbsn}
\begin{table*}[!t]
\small
\centering
\renewcommand{\arraystretch}{1.5}
\begin{tabular}{|m{3.2cm}|p{11.8cm}|}
\hline
\rowcolor[HTML]{F3F3F3} \multicolumn{2}{|l|}{\textbf{Case 1: Modern Human Activity}} \\ \hline
\begin{minipage}{3.2cm}
    \centering
    \vspace{2mm}
    \includegraphics[width=3cm]{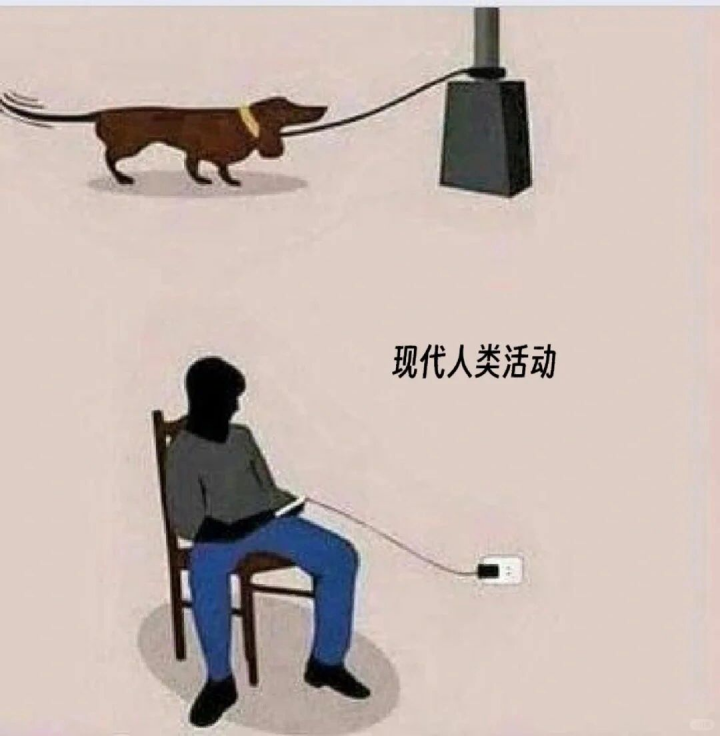} 
    \vspace{2mm}
\end{minipage} & 
\textbf{Text}: 现代人类活动(Modern Human Activity) \newline 
\textbf{Target}: 手机依赖 (Mobile device dependency) \newline 
\textbf{Explanation}: 通过人与狗在相似场景中的对比，讽刺人类对手机及充电设备的过度依赖。反映了科技在提供便利的同时，也带来了生理与精神上的束缚。 \newline 
\textit{(Translation: By contrasting humans and dogs in similar scenarios, this sample satirizes excessive reliance on mobile devices, reflecting how technology limits freedom while providing convenience.)} \\ \hline

\rowcolor[HTML]{F3F3F3} \multicolumn{2}{|l|}{\textbf{Case 2: The Role Model}} \\ \hline
\begin{minipage}{3.2cm}
    \centering
    \vspace{2mm}
    \includegraphics[width=3cm]{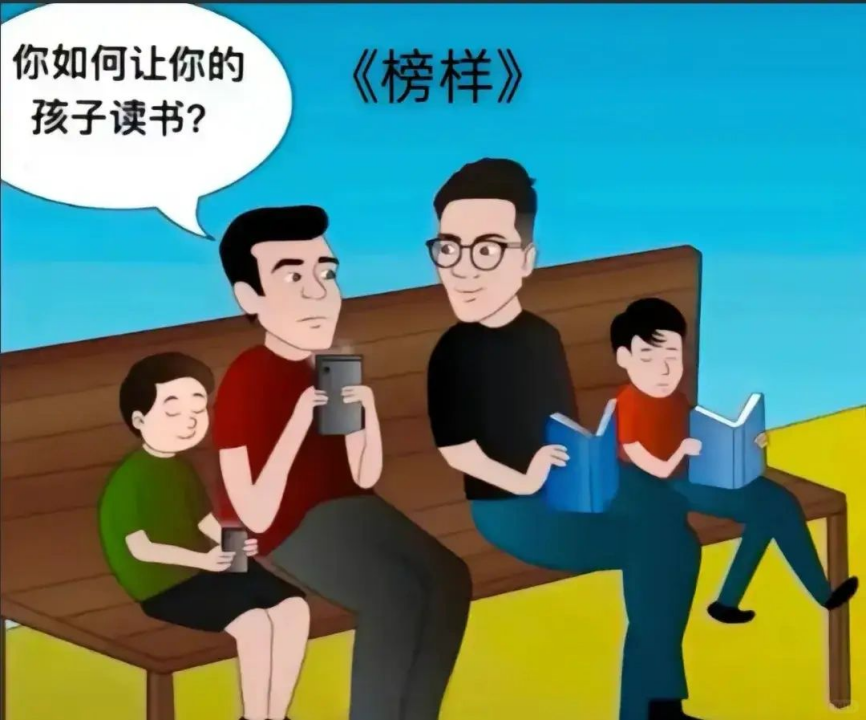}
    \vspace{2mm}
\end{minipage} & 
\textbf{Text}: 《榜样》(《The Role Model》) \newline 
\textbf{Target}: 家长榜样作用缺失 (Lack of parental role modeling) \newline 
\textbf{Explanation}: 左边的家长使用手机，孩子随之模仿，右边的家长在读书，孩子也读书。通过对比揭示了一些家长以身作则不足的问题。 \newline 
\textit{(Translation: This case contrasts two parental behaviors—using phones vs. reading—and the child's subsequent imitation, revealing the failure of parental role modeling in fostering habits.)} \\ \hline

\rowcolor[HTML]{F3F3F3} \multicolumn{2}{|l|}{\textbf{Case 3: Rich vs. Poor}} \\ \hline
\begin{minipage}{3.2cm}
    \centering
    \vspace{2mm}
    \includegraphics[width=3cm]{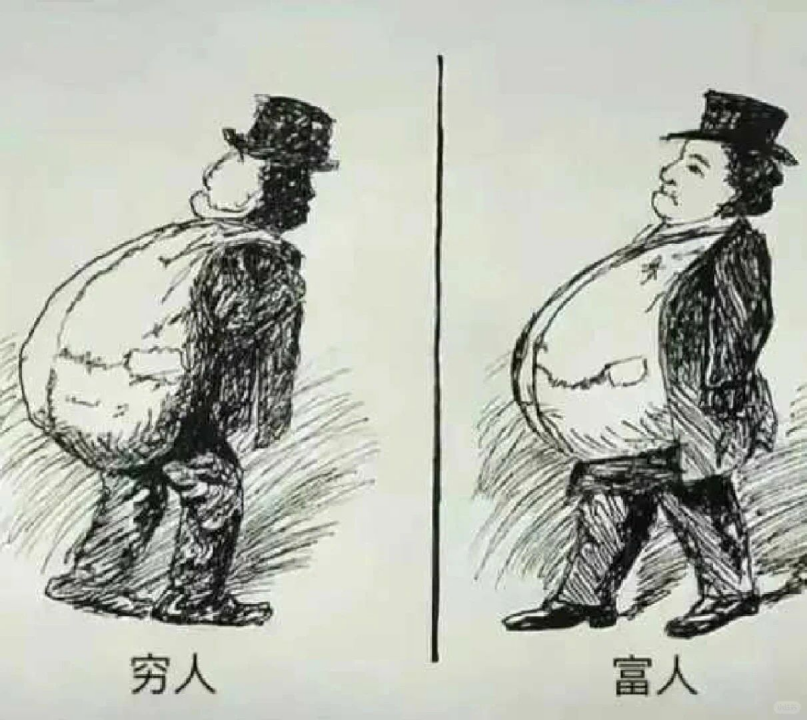}
    \vspace{2mm}
\end{minipage} & 
\textbf{Text}: 穷人 富人(Rich Poor) \newline 
\textbf{Target}: 贫富差距 (Wealth disparity) \newline 
\textbf{Explanation}: 通过对比穷人负重弯曲的脊背与富人鼓胀的肚子，反映富人通过压榨穷人积累财富的现象，批判社会的贫富不均。 \newline 
\textit{(Translation: By comparing the hunched back of the poor carrying heavy loads with the bloated stomach of the rich, it reflects the accumulation of wealth through exploitation and criticizes social inequality.)} \\ \hline
\end{tabular}
\caption{Fine-grained annotations of samples from the CFMS dataset.}
\label{tab:cfms_cases}
\end{table*}
\end{CJK*} 

\subsection{Sarcastic Metaphor Data Construction}
The relationship between metaphor and sarcasm is a cornerstone of linguistic research. Conceptual Metaphor Theory \citep{gibbs2017metaphor} posits that metaphors are fundamental to human cognition, and sarcasm is often built upon metaphoric foundations. Given the high subjectivity of metaphors, the initial Fleiss’ Kappa of our raw annotation was only around 0.15. 

Furthermore, to facilitate cross-linguistic studies, we leveraged a previously collected English metaphor dataset sourced from Twitter archives. This subset, which originally only contained binary metaphor labels, was re-annotated using our triple-level framework to provide a high-quality parallel English reference (Table \ref{tab:en_metaphor}).

\begin{figure}[h]
    \centering
    \includegraphics[width=\columnwidth]{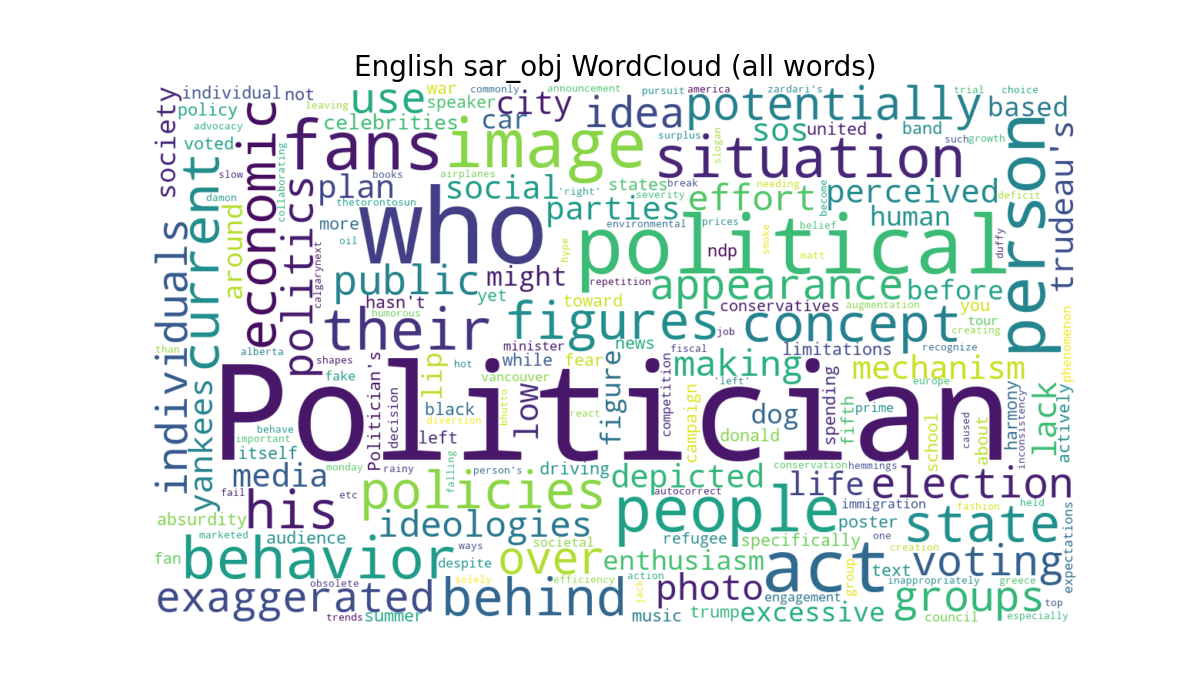} 
    \caption{Word cloud of sarcasm targets in the English parallel subset. It reveals a higher concentration of targets related to public policy and institutional efficiency compared to the Chinese subset.}
    \label{fig:wordcloud_en}
\end{figure}

Comparing the English target word cloud (Figure \ref{fig:wordcloud_en}) with the Chinese one, we observe that English sarcasm targets are more concentrated on \textbf{public policy} and \textbf{institutional efficiency}, whereas Chinese targets frequently focus on \textbf{social etiquette} and \textbf{workplace culture}, reflecting distinct cultural nuances in rhetorical expression.
\begin{table}[ht]
\centering
\small
\begin{tabular}{lcc}
\toprule
\textbf{Type (Chinese)} & \textbf{Positive} & \textbf{Negative} \\ \midrule
Metaphor                & 73                & 127               \\
Sarcasm                 & 71                & 129               \\ \bottomrule
\end{tabular}
\caption{Distribution of the Chinese Sarcasm-Metaphor subset (N=200).}
\label{tab:zh_metaphor}
\end{table}

\begin{table}[ht]
\centering
\small
\begin{tabular}{lcc}
\toprule
\textbf{Type (English)} & \textbf{Positive} & \textbf{Negative} \\ \midrule
Metaphor                & 100               & 100               \\
Sarcasm                 & 100               & 100               \\ \bottomrule
\end{tabular}
\caption{Distribution of the English parallel subset (N=200).}
\label{tab:en_metaphor}
\end{table}

\subsection{Explanation-Driven AI Generation Test Set}
We utilized CFMS annotations to guide \texttt{gemini-3-pro-image-preview} in generating 100 images. Two annotators evaluated whether the generated content achieved the intended sarcastic effect. As shown in Table \ref{tab:ai_gen}, an average of 76\% of the images successfully conveyed the intended sarcasm, confirming the value of CFMS labels as structured instructional signals for controllable generation. Failure cases (e.g., ``Workplace PUA'') reveal current limitations in reifying abstract cultural metaphors.
Qualitative examples of AI-generated sarcastic images can be found in Appendix \ref{subsec:ai_gen}.
\begin{table}[ht]
\centering
\small
\begin{tabular}{lcc}
\toprule
\textbf{Annotator} & \textbf{Sarcastic Effect Met} & \textbf{Ratio (\%)} \\ \midrule
Annotator A        & 70                            & 70.0                \\
Annotator B        & 82                            & 82.0                \\ \midrule
\textbf{Average}   & \textbf{76}                   & \textbf{76.0}       \\ \bottomrule
\end{tabular}
\caption{Evaluation of AI-generated images based on CFMS annotations (N=100).}
\label{tab:ai_gen}
\end{table}

\section{Approach}
\subsection{Task Definition}
CFMS comprises three progressively hierarchical subtasks designed to evaluate model capabilities across identification, localization, and explanation. We define the input as a multimodal pair $x = (I, T)$, where $I$ represents the image and $T$ represents the text.

\textbf{Task 1: Sarcasm Identification.} This task aims to determine whether the input contains sarcastic intent. We define it as a mapping function $f_{\text{det}}$ that maps the multimodal input to a binary label space:
\begin{equation}
f_{\text{det}}: I \times T \rightarrow \{0, 1\}
\end{equation}
where $0$ denotes non-sarcasm and $1$ denotes sarcasm.

\textbf{Evaluation Metrics:} We employ a standard set of classification metrics $\mathcal{M}_{\text{cls}} = \{\text{Accuracy, Precision, Recall, F1}\}$. For the test set $D_{\text{test}}$, the discrepancy between the predicted value $\hat{y} = f_{\text{det}}(x)$ and the ground truth $y$ is calculated.

\textbf{Task 2: Sarcasm Target Recognition.} 
Conditioned on identifying a sarcastic sample ($y=1$), this task aims to localize the target of the sarcasm. It is defined as a mapping from the input to a target set $V$:
\begin{equation}
f_{\text{target}}: (I \times T \mid y=1) \rightarrow t \in V
\end{equation}
where $t$ is the identified specific sarcastic target.

\textbf{Evaluation Metrics:} Given the open-ended nature of generation, we define a semantic consistency function $S(t_{\text{pred}}, t_{\text{gold}}) \in \{0, 1\}$. The accuracy is defined as:
\begin{equation}
\text{Acc}_{\text{target}} = \frac{1}{|D_{\text{pos}}|} \sum_{x \in D_{\text{pos}}} S(f_{\text{target}}(x), t_{\text{gold}})
\end{equation}
where $D_{\text{pos}}$ denotes the subset of sarcastic (positive) instances in the test set, as target recognition is only evaluated on samples with valid sarcastic intent. $|D_{\text{pos}}|$ represents the cardinality of this subset. The consistency function $S$ is implemented via GPT-4o-based semantic matching to account for linguistic variations in generated outputs.

\textbf{Task 3: Sarcasm Explanation Generation.} 
This task requires the model to generate a natural language text elucidating the construction mechanism of the sarcasm. The generation function is defined as follows:
\begin{equation}
f_{\text{exp}}: (I \times T \mid y=1) \rightarrow E = (w_1, w_2, \dots, w_n)
\end{equation}
where $E$ is the sequence of the explanation text.

\textbf{Evaluation Metrics:} We utilize a set of text generation quality metrics $\mathcal{M}_{\text{gen}} = \{\text{BLEU-4, BERTScore}\}$ to measure the semantic similarity between the generated explanation $\hat{E}$ and the human-annotated reference $E_{\text{ref}}$.

\subsection{Policy-Guided Demonstration Selection}
Traditional similarity-based retrieval methods often fall short when handling tasks like sarcasm detection that require deep semantic reasoning. To address this, we propose \textbf{Policy-Guided Demonstration Selection (PGDS)}, a reinforcement-learned strategy that dynamically optimizes exemplar selection to improve fine-grained sarcasm understanding. The comprehensive workflow is outlined in Algorithm \ref{alg:pgds}.

\textbf{Framework Design:} Given a query sample $x_q = (I_q, T_q)$, where $I_q$ is the image and $T_q$ is the text, we first obtain a joint representation using a BGE encoder $f_{\text{BGE}}$ \citep{xiao2023cpack} and a CLIP encoder $f_{\text{CLIP}}$ \citep{radford2021learning}:
\begin{equation}
    h_q = \text{Concat}(f_{\text{BGE}}(T_q), f_{\text{CLIP}}(I_q))
\end{equation}
Based on this representation, we retrieve the top-50 candidate samples from the training set $\mathcal{D}_{\text{train}}$:
\begin{equation}
    \mathcal{C}_q = \{x_i \mid x_i \in \mathcal{D}_{\text{train}}, \text{sim}(h_q, h_i) \in \text{Top-50}\}
\end{equation}
where $\text{sim}(h_q, h_i)$ denotes the cosine similarity.

The core innovation lies in the introduction of a lightweight policy network $\pi_\theta$, parameterized as a two-layer MLP. Taking the joint embedding of the query and candidates as input, it outputs the selection probabilities:
\begin{align}
    p_i &= \pi_\theta(h_q, h_i) \nonumber \\
    &= \text{Softmax}(W_2\sigma(W_1[h_q; h_i] + b_1) + b_2)
\end{align}
where $\sigma$ is the ReLU activation function and $[;]$ denotes vector concatenation. After calculating the weights for each candidate, we employ a probability-based sampling mechanism to select $k$ representative examples: $\mathcal{S}_q = \text{SampleTopK}(\{p_i\}_{i=1}^{50}, k)$. These examples are combined with the query into a structured prompt: $P = \text{Template}(\mathcal{S}_q, x_q)$. The multimodal model $f_{\text{MLLM}}$ then generates a prediction: $\hat{y} = f_{\text{MLLM}}(P)$.

\textbf{Reward Design and Optimization:} Predictions are evaluated through a multi-dimensional reward function:
\begin{equation}
    R = w_1 \cdot R_{\text{format}} + w_2 \cdot R_{\text{cls}} + w_3 \cdot R_{\text{target}} + w_4 \cdot R_{\text{exp}}
\end{equation}
where $R_{\text{format}} \in \{0, 1\}$ assesses the output format compliance, $R_{\text{cls}} \in \{0, 1\}$ evaluates classification accuracy, and $R_{\text{target}}$ and $R_{\text{exp}}$ utilize \text{BERTScore} to measure the semantic quality of the identified target and generated explanation, respectively. The weights $w_i$ are tuned via the validation set such that $\sum w_i = 1$.

The policy network is optimized using the REINFORCE algorithm \citep{williams1992simple}, with the objective function: $J(\theta) = \mathbb{E}_{\mathcal{S}_q \sim \pi_\theta}[R(\mathcal{S}_q)]$. To reduce variance, gradient estimation incorporates a moving average baseline $b_t$:
\begin{equation}
    \nabla_\theta J(\theta) \approx \frac{1}{N} \sum_{i=1}^N (R_i - b_t) \nabla_\theta \log \pi_\theta(\mathcal{S}_q^{(i)} \mid h_q^{(i)})
\end{equation}
where $b_t = \gamma b_{t-1} + (1-\gamma)R_{t-1}$ is the exponential moving average with $\gamma=0.9$.

Compared to static retrieval, our dynamic strategy demonstrates advantages in complex sarcastic scenarios. Particularly for samples involving metaphors, the policy network identifies examples that may have low surface similarity but are highly relevant in deep semantic structure. This adaptability is crucial for handling the diverse sarcastic expressions prevalent in Chinese social media.

\begin{algorithm}[t]
\small
\caption{PGDS: Policy-Guided Demonstration Selection}
\label{alg:pgds}
\begin{algorithmic}[1]
\REQUIRE Training set $\mathcal{D}_{\text{train}}$, Query $x_q$, MLLM $f_{\text{MLLM}}$, Policy $\pi_\theta$, Exemplar size $k$
\ENSURE Selected demonstrations $\mathcal{S}_q$ for inference
\STATE $h_q \gets \text{Encode}(x_q)$ \COMMENT{BGE for text, CLIP for image}
\STATE $\mathcal{C}_q \gets \text{TopK}(\{\text{sim}(h_q, h_i) \mid x_i \in \mathcal{D}_{\text{train}}\}, 50)$
\FOR{each candidate $x_i \in \mathcal{C}_q$}
    \STATE $p_i \gets \pi_\theta(h_q, h_i)$ \COMMENT{Compute selection probability}
\ENDFOR
\STATE $\mathcal{S}_q \gets \text{Sample}(\mathcal{C}_q, \{p_i\}, k)$ \COMMENT{Probabilistic sampling}
\STATE $P \gets \text{Template}(\mathcal{S}_q, x_q)$ \COMMENT{Construct structured prompt}
\STATE $\hat{y} \gets f_{\text{MLLM}}(P)$ \COMMENT{Generate interpretation}
\STATE $R \gets \text{CalculateReward}(\hat{y}, y_{gold})$ \COMMENT{Format, Cls, BERTScore}
\IF{in training phase}
    \STATE $b_t \gets \gamma b_{t-1} + (1-\gamma)R$ \COMMENT{Update baseline}
    \STATE $\theta \gets \theta + \alpha (R - b_t) \nabla_\theta \log \pi_\theta(\mathcal{S}_q \mid h_q)$ \COMMENT{Policy gradient update}
\ENDIF
\STATE \RETURN $\mathcal{S}_q$
\end{algorithmic}
\end{algorithm}
\section{Experiments}
\begin{table*}[htb]
\centering
\small
\setlength{\tabcolsep}{4.5pt}
\begin{tabular}{ll cccc ccc}
\toprule
\textbf{Model} & \textbf{Method} & \textbf{Acc} $\uparrow$ & \textbf{P} & \textbf{R} & \textbf{F1} $\uparrow$ & \textbf{Target Acc} $\uparrow$ & \textbf{BLEU-4 ($\%$)} & \textbf{BERTScore ($\%$)} \\
\midrule
\multicolumn{9}{c}{\textbf{\textsc{Closed-Source}}} \\
\midrule
\multirow{3}{*}{GPT-4o}
& Zero-shot & 70.24 & 60.67 & 96.30 & 74.44 & 71.03 & 8.07 & 72.81 \\
& Random 1-shot & 76.36 & 70.08 & 86.41 & 77.39 & 61.17 & 8.39 & 72.59 \\
& RAG 1-shot & 77.62 & 68.20 & 94.18 & 79.10 & 73.02 & \textbf{10.60} & \textbf{73.84} \\
\midrule
\multirow{3}{*}{Gemini-2.5-Flash}
& Zero-shot & 71.19 & 60.31 & 95.24 & 74.84 & \textbf{82.54} & 6.30 & 72.41 \\
& Random 1-shot & 75.71 & 67.61 & 88.36 & 76.61 & 76.19 & 6.88 & 72.55 \\
& RAG 1-shot & 75.00 & 65.79 & 92.59 & 76.92 & 80.42 & 7.34 & 73.03 \\
\midrule
\multicolumn{9}{c}{\textbf{\textsc{Open-Source}}} \\
\midrule
\multirow{4}{*}{Qwen2.5-VL-7B-Instruct}
& Zero-shot & 70.71 & 61.64 & 73.54 & 69.33 & 32.54 & 7.29 & 72.04 \\
& Random 1-shot & 71.25 & 64.54 & 80.16 & 71.50 & 38.62 & 8.24 & 71.83 \\
& RAG 1-shot & 76.88 & 71.08 & 81.94 & 76.13 & 45.67 & 8.18 & 71.92 \\
& PGDS (Ours) & 78.01 & 76.92 & 75.76 & 76.34 & 48.68 & 8.10 & 71.52 \\
\midrule
\multirow{4}{*}{InternVL2.5-8B}
& Zero-shot & 68.48 & 59.67 & 93.72 & 72.91 & 40.47 & 6.77 & 70.72 \\
& Random 1-shot & 69.29 & 60.10 & 94.44 & 73.46 & 48.15 & 8.33 & 72.31 \\
& RAG 1-shot & 71.05 & 62.11 & 96.15 & 75.47 & 46.15 & 8.58 & 71.69 \\
& PGDS (Ours) & 74.76 & 66.27 & 89.42 & 76.13 & 50.89 & 8.76 & 72.30 \\
\bottomrule
\end{tabular}
\caption{Main results on the CFMS dataset (excluding LoRA fine-tuning). \textbf{Zero-shot}, \textbf{Random 1-shot}, \textbf{RAG 1-shot} are baselines; \textbf{PGDS (Ours)} is our proposed dynamic demonstration selection strategy. Best results per column (across all models/methods) are in \textbf{bold}. "-" indicates no corresponding experimental results.}
\label{tab:main_results}
\end{table*}

\begin{table*}[htb]
\centering
\small
\setlength{\tabcolsep}{4.5pt}
\begin{tabular}{ll cccc ccc}
\toprule
\textbf{Model} & \textbf{Method} & \textbf{Acc} $\uparrow$ & \textbf{P} & \textbf{R} & \textbf{F1} $\uparrow$ & \textbf{Target Acc} $\uparrow$ & \textbf{BLEU-4 ($\%$)} & \textbf{BERTScore ($\%$)} \\
\midrule
Qwen2.5-VL-7B-Instruct & LoRA FT & 82.30 & 81.22 & 78.61 & 79.89 & 49.73 & 7.51 & 71.24 \\
InternVL2.5-8B & LoRA FT & \textbf{83.25} & 79.70 & 83.96 & \textbf{81.77} & 53.48 & 9.35 & 72.36 \\
LLaVA-1.5-7B & LoRA FT & 80.95 & \textbf{84.71} & 70.37 & 76.88 & 36.51 & 5.71 & 69.32 \\
\bottomrule
\end{tabular}
\caption{Experimental results of LoRA fine-tuning (LoRA FT) on the CFMS dataset. Best results per column are in \textbf{bold}.}
\label{tab:lora_ft_results}
\end{table*}

\begin{table*}[htb]
\centering
\small
\setlength{\tabcolsep}{4.5pt}
\begin{tabular}{l | cc | cc | c || cc | cc | c}
\toprule
\multirow{3}{*}{\textbf{Model}} & \multicolumn{5}{c||}{\textbf{Chinese Subset (N=200)}} & \multicolumn{5}{c}{\textbf{English Subset (N=200)}} \\
\cmidrule(lr){2-6} \cmidrule(lr){7-11}
& \multicolumn{2}{c}{Sarcasm} & \multicolumn{2}{c}{Metaphor} & \multirow{2}{*}{$\Delta_{\text{F1}}$} & \multicolumn{2}{c}{Sarcasm} & \multicolumn{2}{c}{Metaphor} & \multirow{2}{*}{$\Delta_{\text{F1}}$} \\
& Acc. & F1 & Acc. & F1 & & Acc. & F1 & Acc. & F1 & \\
\midrule
GPT-4o & 77.00 & 75.53 & 41.50 & 55.51 & -20.02 & \textbf{79.00} & \textbf{81.25} & 58.00 & 69.78 & -11.47 \\
Gemini-2.5-Flash & 77.00 & 75.27 & 47.50 & 57.49 & -17.78 & 72.00 & 74.31 & \textbf{64.50} & \textbf{72.58} & \textbf{-1.73} \\
Qwen2.5-VL-7B-Instruct & \textbf{84.50} & 78.77 & 60.50 & 59.49 & -19.28 & 67.00 & 66.67 & 55.00 & 66.17 & -0.50 \\
InternVL2.5-8B & 81.50 & \textbf{79.10} & \textbf{63.75} & \textbf{64.02} & -15.08 & 70.00 & 63.41 & 54.00 & 58.68 & -4.73 \\
LLaVA-1.5-7B & 35.50 & 52.40 & 36.50 & 53.48 & +1.08 & 57.00 & 29.51 & 44.50 & 17.78 & -11.73 \\
\bottomrule
\end{tabular}
\caption{Binary classification performance (\%) comparison between Sarcasm and Metaphor identification across Chinese and English subsets. $\Delta_{\text{F1}}$ denotes the F1-score drop from sarcasm to metaphor.}
\label{tab:metaphor_results}
\end{table*}

\subsection{Experimental Setup}

\textbf{Baselines:} We evaluate a diverse set of state-of-the-art MLLMs, including: (1) \textbf{Closed-source models:} GPT-4o \cite{hurst2024gpt} and Gemini-2.5-Flash \cite{comanici2025gemini}. (2) \textbf{Open-source models:} Qwen2.5-VL-7B-Instruct \cite{bai2025qwen2}, InternVL2.5-8B \cite{chen2025expandingperformanceboundariesopensource}, and LLaVA-1.5-7B \cite{NEURIPS2023_6dcf277e}. Note that LLaVA-1.5-7B is excluded from 1-shot experiments as it does not natively support interleaved multi-image inputs.

\textbf{Learning Paradigms:}
To explore different adaptation strategies, we implement two paradigms:
\begin{itemize}
    \item \textbf{Parameter-Efficient Fine-Tuning (PEFT):} We employ LoRA \cite{hu2022lora} to adapt models by introducing low-rank matrices $A$ and $B$ to approximate the weight update:
    \begin{equation}
    h = Wx + BAx
    \end{equation}
    This allows the model to learn sarcastic patterns with minimal computational cost.
    
    \item \textbf{Retrieval-Based In-Context Learning (ICL):} We adopt an ICL framework that retrieves the most similar samples $s_i$ from $D_{\text{train}}$ to construct an augmented prompt:
    \begin{equation}
    P = [s_1 \oplus \dots \oplus s_k \oplus x_q]
    \end{equation}
    The similarity is determined by the cosine distance of textual features:
    \begin{equation}
    \text{sim}(x_q, x_i) = \frac{\text{Enc}(T_q) \cdot \text{Enc}(T_i)}{\|\text{Enc}(T_q)\| \cdot \|\text{Enc}(T_i)\|}
    \end{equation}
\end{itemize}

\textbf{Implementation Details:} For LoRA, we set $r=16, \alpha=32$, targeting all linear modules with a $5 \times 10^{-5}$ learning rate over 10 epochs. For ICL, we employ a multimodal retrieval strategy where \texttt{bge-large-zh-v1.5} is used for textual semantic encoding and \texttt{clip-vit-base-patch32} is used for image feature extraction. These multimodal embeddings are indexed using FAISS to facilitate efficient and accurate demonstration retrieval.

\textbf{Evaluation Metrics:} We report Accuracy (Acc), Precision (P), Recall (R), and F1-score for identification. For target recognition, GPT-4o serves as a semantic judge. Explanation generation is evaluated via BLEU-4 and BERTScore-F1.

\subsection{Main Results and Analysis}
Table \ref{tab:main_results} and Table \ref{tab:lora_ft_results} summarize model performance across different paradigms.

\textbf{Performance of ICL Strategies:} As shown in Table \ref{tab:main_results}, 1-shot learning generally improves performance over zero-shot. While Random 1-shot provides a baseline, RAG 1-shot further leverages semantic similarity to enhance reasoning. However, closed-source models like Gemini-2.5-Flash show high sensitivity to demonstration quality, where Random 1-shot leads to a drop in Target Acc (from 82.54\% to 76.19\%).

\textbf{The Superiority and Efficiency of PGDS:} 
Our proposed \textbf{PGDS} consistently outperforms Random and RAG-based 1-shot baselines. For InternVL2.5-8B, PGDS improves Acc to 74.76\% and Target Acc to 50.89\%. Notably, PGDS achieves these gains without the heavy computational cost of parameter updates required by LoRA FT. While LoRA FT (Table \ref{tab:lora_ft_results}) provides the highest upper bound for classification (e.g., InternVL2.5-8B reaching 83.25\% Acc), PGDS offers a cost-effective alternative that bridges the gap between zero-shot and full fine-tuning via policy-driven context engineering.

\textbf{Target Recognition Challenges:} Despite strong identification results, target recognition remains a bottleneck. Even fine-tuned models (max 53.48\% Target Acc) fail to reach the zero-shot performance of Gemini-2.5-Flash (82.54\%). This suggests that precise entity localization in sarcasm requires deeper cross-modal reasoning that exceeds the capacity of current small-scale open-source model fine-tuning.

\subsection{Probing Deep Semantics: Metaphor vs. Sarcasm}
\label{sec:metaphor_sarcasm}

To evaluate the boundaries of deep semantic reasoning, we conduct a comparative study on a subset of 200 high-consistency ``Metaphor-Sarcasm'' pairs in Chinese and an equal-sized English parallel subset. Results are summarized in Table \ref{tab:metaphor_results}.

\textbf{Cross-lingual Consistency of Metaphorical Difficulty:} 
Metaphor identification is consistently more challenging than sarcasm detection across languages. In the Chinese subset, the performance gap ($\Delta_{\text{F1}}$) for mainstream models (excluding LLaVA) ranges from 15\% to 20\%. A similar trend is observed in the English subset, where GPT-4o shows an 11.47\% drop. This quantifiably validates \citet{gibbs2017metaphor}'s theory that metaphor interpretation necessitates complex cross-domain mapping and deeper cognitive reasoning than sentiment-driven sarcasm.

\textbf{Cultural Dependency of Model Performance:} 
Model performance is highly correlated with the cultural attributes of their pre-training data. In the \textbf{Chinese context}, the open-source InternVL2.5-8B achieves the best metaphor F1 (64.02\%), significantly outperforming GPT-4o (55.51\%), suggesting that localized multimodal alignment enhances understanding of regional rhetoric. In the \textbf{English context}, GPT-4o maintains its lead in sarcasm detection (81.25\%), while Gemini-2.5-Flash shows superior adaptability in metaphors with a minimal gap ($\Delta_{\text{F1}}=1.73\%$), indicating robust reasoning in English semantics.

\textbf{Limitations of Shallow Alignment:} 
LLaVA-1.5-7B fails to demonstrate effective discrimination in both subsets. Its near-zero $\Delta_{\text{F1}}$ in the Chinese task indicates a reliance on shallow surface features rather than mastering deep semantic alignment, resulting in near-random predictions.

\section{Conclusion}
We have presented \textbf{CFMS}, a fine-grained Chinese multimodal sarcasm benchmark with triple-annotation. To enhance model performance, we propose \textbf{PGDS }, a reinforcement learning-based strategy for dynamic demonstration selection, which effectively narrows the gap between zero-shot inference and fine-tuning. Our systematic evaluation reveals that while LoRA-tuned models achieve high classification accuracy, capturing high-order metaphorical logic and assessing AI-generated sarcastic content remain significant challenges. CFMS provides a robust foundation for future research in interpretable affective computing and complex semantic alignment.

\section*{Limitations}
Despite its contributions, this work has several limitations. First, CFMS is built on Chinese social media data; in the future, we could consider annotating more diverse data to expand its cultural and linguistic scale. Second, while PGDS  improves performance, it still faces challenges in high-order metaphorical reasoning and involves higher computational costs than static ICL. Finally, the inherent subjectivity of sarcasm means that individual cognitive biases in interpretation remain unavoidable.

\section*{Ethics Statement}
This research adheres to the ACL Ethics Policy. All data, comprising Chinese social media samples and English Twitter archives, are sourced from public platforms. We have strictly anonymized all personal identifiers and ensured that visual content contains no identifiable individuals to protect user privacy. Annotations were performed by students who received fair compensation above local labor standards. We employed a human-in-the-loop approach to verify and correct all LLM-generated pre-annotations, mitigating potential algorithmic biases. The CFMS benchmark and PGDS strategy are intended for beneficial research in explainable AI and creative synthesis; we do not condone their use for generating harmful, deceptive, or discriminatory content.

\bibliography{custom}

@inproceedings{cai2019multi,
  title={Multi-modal sarcasm detection in {Twitter} with hierarchical fusion model},
  author={Cai, Yitao and Cai, Huiyu and Wan, Xiaojun},
  booktitle={Proceedings of the 57th Annual Meeting of the Association for Computational Linguistics},
  pages={2506--2515},
  year={2019}
}

@inproceedings{castro2019towards,
  title={Towards multimodal sarcasm detection (an obviously perfect paper)},
  author={Castro, Santiago and Hazarika, Devamanyu and P{\'e}rez-Rosas, Ver{\'o}nica and Zimmermann, Roger and Mihalcea, Rada and Poria, Soujanya},
  booktitle={Proceedings of the 57th Annual Meeting of the Association for Computational Linguistics},
  pages={4619--4629},
  year={2019}
}

@inproceedings{lu2023seeing,
  title={Seeing is not always believing: Benchmarking human and model perception of {AI}-generated images},
  author={Lu, Zeyu and Huang, Di and Bai, Lei and Qu, Jingjing and Wu, Chengyue and Liu, Xiaoxian and Ouyang, Wanli},
  booktitle={Advances in Neural Information Processing Systems},
  volume={36},
  pages={25435--25447},
  year={2023}
}

@article{farabi2024survey,
  title={A survey of multimodal sarcasm detection},
  author={Farabi, Sahar and Ranasinghe, Tharindu and Kanojia, Diptesh and Kong, Ying and Zampieri, Marcos},
  journal={arXiv preprint arXiv:2410.18882},
  year={2024}
}

@book{gibbs2017metaphor,
  title={Metaphor wars: Conceptual metaphors in human life},
  author={Gibbs, Raymond W},
  publisher={Cambridge University Press},
  year={2017}
}

@inproceedings{hu2022lora,
  title={{LoRA}: Low-rank adaptation of large language models},
  author={Hu, Edward J and Shen, Yelong and Wallis, Phillip and Allen-Zhu, Zeyuan and Li, Yuanzhi and Wang, Shean and Chen, Weizhu},
  booktitle={International Conference on Learning Representations},
  year={2022}
}

@article{konstantinidou2025navigating,
  title={Navigating the challenges of {AI}-generated image detection in the wild: What truly matters?},
  author={Konstantinidou, Despina and Karageorgiou, Dimitrios and Koutlis, Christos and Papadopoulou, Olga and Schinas, Emmanouil and Papadopoulos, Symeon},
  journal={arXiv preprint arXiv:2507.10236},
  year={2025}
}

@inproceedings{liang2022multi,
  title={Multi-modal sarcasm detection via cross-modal graph convolutional network},
  author={Liang, Bin and Lou, Chenwuei and Li, Xin and Yang, Min and Gui, Lin and He, Yulan and Xu, Ruifeng},
  booktitle={Proceedings of the 60th Annual Meeting of the Association for Computational Linguistics (Volume 1: Long Papers)},
  pages={1767--1777},
  year={2022}
}

@inproceedings{qin2023multi,
  title={{MMSD2.0}: Towards a reliable multi-modal sarcasm detection system},
  author={Qin, Libo and Huang, Shijue and Chen, Qiguang and Cai, Chenran and Zhang, Yudi and Liang, Bin and Che, Wanxiang and Xu, Ruifeng},
  booktitle={Findings of the Association for Computational Linguistics: ACL 2023},
  pages={10834--10845},
  year={2023}
}

@inproceedings{schifanella2016multimodal,
  title={Detecting sarcasm in multimodal social platforms},
  author={Schifanella, Rossano and Juan, Paloma De and Tetreault, Joel R and Cao, Liangliang},
  booktitle={Proceedings of the 24th ACM International Conference on Multimedia},
  year={2016}
}

@inproceedings{yue2024sarcnet,
  title={{SarcNet}: A multilingual multimodal sarcasm detection dataset},
  author={Yue, Tianqi and Shi, Xiaochen and Mao, Rui and Hu, Zheheng and Cambria, Erik},
  booktitle={Proceedings of the 2024 Joint International Conference on Computational Linguistics, Language Resources and Evaluation (LREC-COLING 2024)},
  pages={14325--14335},
  year={2024}
}

@article{van2005puns,
  title={Puns, relevance and appreciation in advertisements},
  author={Van Mulken, Margot and Van Enschot-van Dijk, Renske and Hoeken, Hans},
  journal={Journal of pragmatics},
  volume={37},
  number={5},
  pages={707--721},
  year={2005},
  publisher={Elsevier}
}

@article{oppenlaender2024taxonomy,
  title={A taxonomy of prompt modifiers for text-to-image generation},
  author={Oppenlaender, Jonas},
  journal={Behaviour \& Information Technology},
  volume={43},
  number={15},
  pages={3763--3776},
  year={2024},
  publisher={Taylor \& Francis}
}

@book{munro2021human,
  title={Human-in-the-loop machine learning: Active learning and annotation for human-centered AI},
  author={Munro, Robert},
  year={2021},
  publisher={Manning Publications}
}

@article{wang2021want,
  title={Want to reduce labeling cost? GPT-3 can help},
  author={Wang, Shuohang and Liu, Yang and Xu, Yichong and Zhu, Chenguang and Zeng, Michael},
  journal={arXiv preprint arXiv:2108.13487},
  year={2021}
}

@article{gilardi2023chatgpt,
  title={ChatGPT outperforms crowd workers for text-annotation tasks},
  author={Gilardi, Fabrizio and Alizadeh, Meysam and Kubli, Ma{\"e}l},
  journal={Proceedings of the National Academy of Sciences},
  volume={120},
  number={30},
  pages={e2305016120},
  year={2023},
  publisher={National Academy of Sciences}
}

@inproceedings{xiao2023cpack,
  title={C-pack: Packed resources for general chinese embeddings},
  author={Xiao, Shitao and Liu, Zheng and Zhang, Peitian and Muennighoff, Niklas and Lian, Defu and Nie, Jian-Yun},
  booktitle={Proceedings of the 47th international ACM SIGIR conference on research and development in information retrieval},
  pages={641--649},
  year={2024}
}

@inproceedings{radford2021learning,
  title={Learning transferable visual models from natural language supervision},
  author={Radford, Alec and Kim, Jong Wook and Hallacy, Chris and Ramesh, Aditya and Goh, Gabriel and Agarwal, Sandhini and Sastry, Girish and Askell, Amanda and Mishkin, Pamela and Clark, Jack and others},
  booktitle={International Conference on Machine Learning},
  pages={8748--8763},
  year={2021},
  organization={PMLR}
}

@article{williams1992simple,
  title={Simple statistical gradient-following algorithms for connectionist reinforcement learning},
  author={Williams, Ronald J},
  journal={Machine learning},
  volume={8},
  number={3},
  pages={229--256},
  year={1992},
  publisher={Springer}
}

@article{camp2012sarcasm,
  title={Sarcasm, pretense, and the semantics/pragmatics distinction},
  author={Camp, Elisabeth},
  journal={No{\^u}s},
  volume={46},
  number={4},
  pages={587--634},
  year={2012},
  publisher={Wiley Online Library}
}

@misc{chen2025expandingperformanceboundariesopensource,
      title={Expanding Performance Boundaries of Open-Source Multimodal Models with Model, Data, and Test-Time Scaling}, 
      author={Zhe Chen and Weiyun Wang and Yue Cao and Yangzhou Liu and Zhangwei Gao and Erfei Cui and Jinguo Zhu and Shenglong Ye and Hao Tian and Zhaoyang Liu and Lixin Gu and Xuehui Wang and Qingyun Li and Yimin Ren and Zixuan Chen and Jiapeng Luo and Jiahao Wang and Tan Jiang and Bo Wang and Conghui He and Botian Shi and Xingcheng Zhang and Han Lv and Yi Wang and Wenqi Shao and Pei Chu and Zhongying Tu and Tong He and Zhiyong Wu and Huipeng Deng and Jiaye Ge and Kai Chen and Kaipeng Zhang and Limin Wang and Min Dou and Lewei Lu and Xizhou Zhu and Tong Lu and Dahua Lin and Yu Qiao and Jifeng Dai and Wenhai Wang},
      year={2025},
      eprint={2412.05271},
      archivePrefix={arXiv},
      primaryClass={cs.CV},
      url={https://arxiv.org/abs/2412.05271}, 
}

@inproceedings{NEURIPS2023_6dcf277e,
 author = {Liu, Haotian and Li, Chunyuan and Wu, Qingyang and Lee, Yong Jae},
 booktitle = {Advances in Neural Information Processing Systems},
 editor = {A. Oh and T. Naumann and A. Globerson and K. Saenko and M. Hardt and S. Levine},
 pages = {34892--34916},
 publisher = {Curran Associates, Inc.},
 title = {Visual Instruction Tuning},
 url = {https://proceedings.neurips.cc/paper_files/paper/2023/file/6dcf277ea32ce3288914faf369fe6de0-Paper-Conference.pdf},
 volume = {36},
 year = {2023}
}

@article{bai2025qwen2,
  title={Qwen2. 5-vl technical report},
  author={Bai, Shuai and Chen, Keqin and Liu, Xuejing and Wang, Jialin and Ge, Wenbin and Song, Sibo and Dang, Kai and Wang, Peng and Wang, Shijie and Tang, Jun and others},
  journal={arXiv preprint arXiv:2502.13923},
  year={2025}
}

@article{hurst2024gpt,
  title={Gpt-4o system card},
  author={Hurst, Aaron and Lerer, Adam and Goucher, Adam P and Perelman, Adam and Ramesh, Aditya and Clark, Aidan and Ostrow, AJ and Welihinda, Akila and Hayes, Alan and Radford, Alec and others},
  journal={arXiv preprint arXiv:2410.21276},
  year={2024}
}

@article{comanici2025gemini,
  title={Gemini 2.5: Pushing the frontier with advanced reasoning, multimodality, long context, and next generation agentic capabilities},
  author={Comanici, Gheorghe and Bieber, Eric and Schaekermann, Mike and Pasupat, Ice and Sachdeva, Noveen and Dhillon, Inderjit and Blistein, Marcel and Ram, Ori and Zhang, Dan and Rosen, Evan and others},
  journal={arXiv preprint arXiv:2507.06261},
  year={2025}
}

\appendix
\section{Appendix}
\label{sec:appendix}

\begin{CJK*}{UTF8}{gbsn} 

\subsection{AI-Generated Sarcastic Content}
\label{subsec:ai_gen}
\begin{figure}[!ht]
\centering
\includegraphics[width=0.85\linewidth]{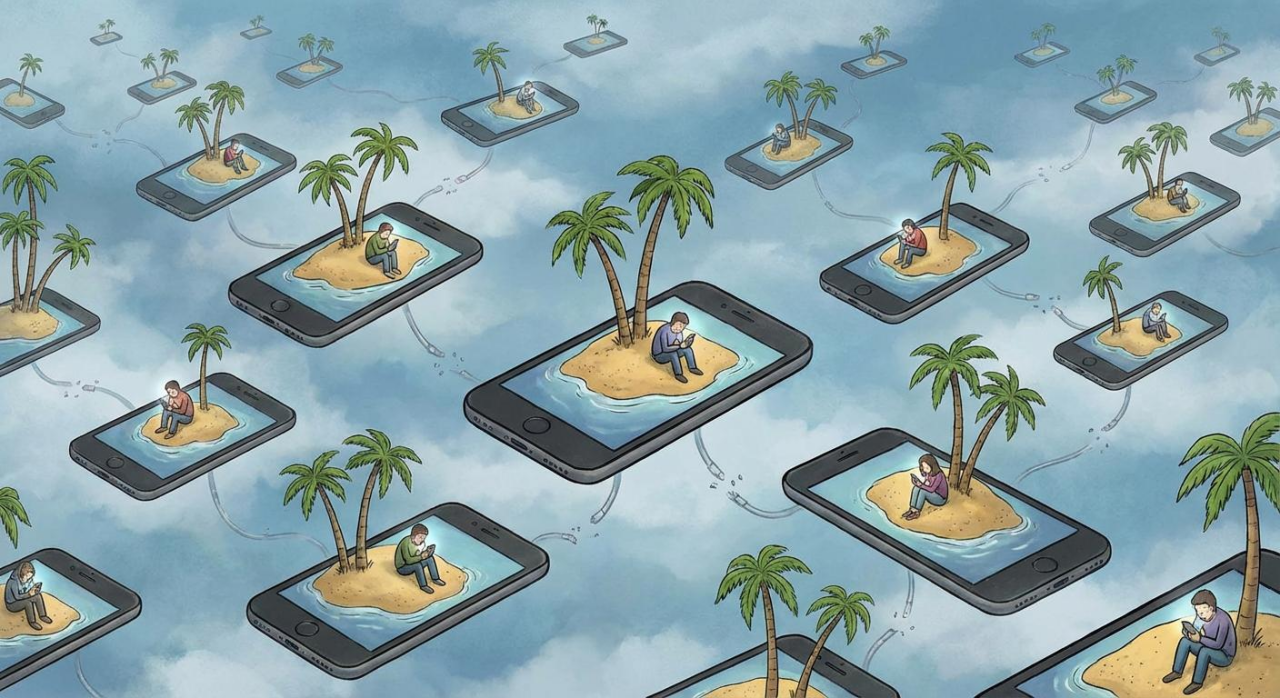} 
\caption{Examples of AI-generated images under sarcastic prompts.}
\label{fig:ai_gen}
\end{figure}

Figure \ref{fig:ai_gen} illustrates sarcastic images generated by different MLLMs. We observe that while models can generate visual associations, they struggle to construct deep ``sarcastic logic'' involving semantic conflicts between text and visual features.

\subsection{Prompt Templates}
\label{subsec:app_prompts}

\subsubsection{GPT-4o Pre-annotation Prompt}
\label{subsubsec:gpt_prompt}
The following prompt employs a Chain-of-Thought (CoT) and critical thinking framework to assist in high-quality initial labeling.

\begin{quote}
\small
\textbf{Original Chinese Prompt:}\\
请严格遵循批判性思维原则分析 <文本-图片对>，按步骤评估讽刺可能性：\\
图片所配文本为：\{文本\} (可能为空) \\
\textbf{【分析框架】} \\
1. \textbf{显性要素记录}：文本字面含义；图片客观描述；文化语境标注。 \\
2. \textbf{矛盾检测}（需满足至少两项）：图文语义冲突；情感基调错位；符号系统反常；存在双层含义线索。 \\
3. \textbf{反假设验证}：替代解释测试；作者意图考量；受众认知调查。 \\
\textbf{【输出规范】} \\
确认讽刺存在的必要条件：①存在可验证的语义对立；②符合常规讽刺表达范式；③排除字面解释合理性。 \\
否定讽刺的充分条件：①图文存在自洽的非讽刺解释；②矛盾强度低于文化认知阈值；③缺乏双层意义证据。 \\
注意：不要把好的现象强行解释为讽刺，不确定时统一判为无讽刺。输出对象需简短。 \\
\textbf{【结构化输出】} \\
<result> <讽刺对象>...</讽刺对象> <讽刺解释>...</讽刺解释> </result>

\vspace{2mm}
\textbf{English Translation:}\\
\textit{Please strictly follow critical thinking principles to analyze the <image-text pair> and evaluate the probability of sarcasm step-by-step: \\
The caption provided is: \{text\} (may be empty) \\
\textbf{[Analysis Framework]} \\
1. \textbf{Explicit Element Recording}: Literal meaning of text; objective description of image; cultural context tagging. \\
2. \textbf{Contradiction Detection} (At least two required): Semantic conflict; emotional misalignment; anomalous use of symbols; clues of double meaning. \\
3. \textbf{Counter-hypothesis Validation}: Alternative explanation testing; author intent; audience perception survey. \\
\textbf{[Output Norms]} \\
Necessary conditions: (1) Verifiable semantic opposition; (2) Alignment with sarcastic paradigms; (3) Exclusion of literal plausibility. \\
Sufficient conditions to negate: (1) Self-consistent non-sarcastic interpretation; (2) Contradiction below cognitive thresholds; (3) Lack of double meaning evidence. \\
Note: Do not interpret positive phenomena as sarcasm. If uncertain, label as non-sarcastic. \\
\textbf{[Structured Output]} \\
<result> <target>...</target> <explanation>...</explanation> </result>}
\end{quote}

\subsubsection{Evaluation and Generation Prompts}
\label{subsubsec:eval_prompts}
Table \ref{tab:app_eval_prompts} provides the prompts for zero-shot detection, few-shot ICL, and AI-guided generation.

\begin{table*}[!t]
\small
\centering
\renewcommand{\arraystretch}{1.5}
\begin{tabular}{|l|p{6cm}|p{6.5cm}|}
\hline
\rowcolor[HTML]{F3F3F3} \textbf{Task} & \textbf{Original Chinese Prompt} & \textbf{English Translation} \\ \hline
\textbf{Zero-shot} & 给你一张图片，图片配文为:\{text\}。分析该图文对是否含有讽刺，并给出讽刺对象和解释... & Given an image with caption: \{text\}. Analyze whether this pair contains sarcasm, and provide the target and explanation... \\ \hline
\textbf{Few-shot} & Example: 输入：配文:\{text\}; 图片:\{img\}. 输出：是否讽刺:\{y/n\}; 讽刺对象:\{target\}; 讽刺解释:\{exp\}. \newline Test: 给你一张图片,分析该图片是否含有讽刺... & Example: Input: \{text/img\}; Output: \{y/n; target; exp\}. \newline Test: Given an image, analyze whether it contains sarcasm and provide the target/explanation. \\ \hline
\textbf{AI Gen} & 请生成一幅讽刺图片，图片描述是：\{描述\}；讽刺对象为：\{对象\}；讽刺解释为：\{解释\}。 & Please generate a sarcastic image. Description: \{desc\}; Target: \{target\}; Explanation: \{exp\}. \\ \hline
\end{tabular}
\caption{Standardized prompts for model evaluation and generation tasks.}
\label{tab:app_eval_prompts}
\end{table*}

\subsection{Annotation Interface}
\label{subsec:app_ui}
\begin{figure}[!ht]
\centering
\includegraphics[width=\linewidth]{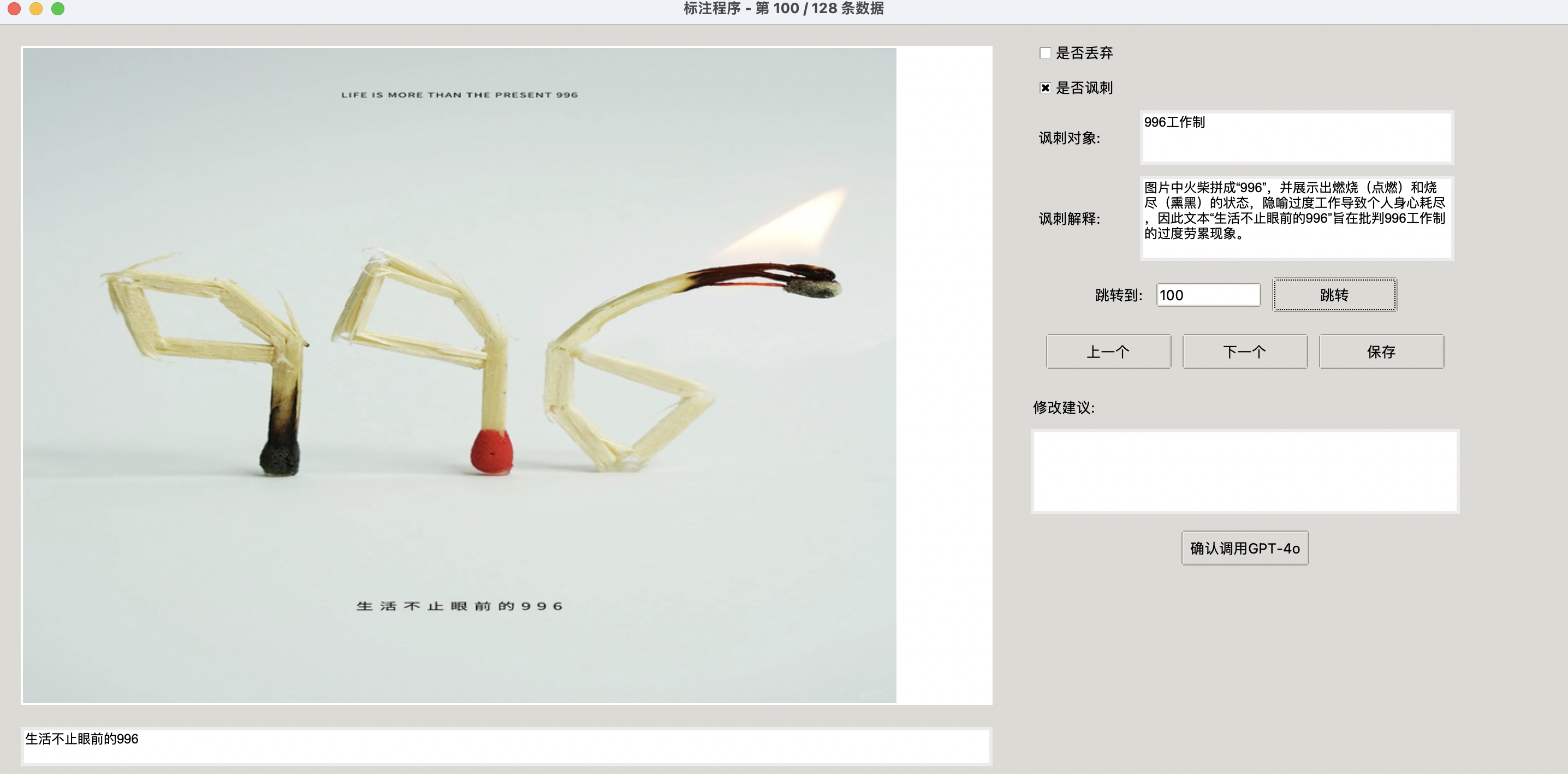}
\caption{Snapshot of the custom-developed CFMS annotation platform.}
\label{fig:ui_shot}
\end{figure}

Figure \ref{fig:ui_shot} displays the custom Web-based annotation platform, which supports synchronized image-text viewing, target selection, and multi-stage verification.

\end{CJK*} 

\end{document}